\def\year{2021}\relax
\documentclass[letterpaper]{article} 
\usepackage{aaai21}  
\usepackage{times}  
\usepackage{helvet} 
\usepackage{courier}  
\usepackage[hyphens]{url}  
\usepackage{graphicx} 
\usepackage{natbib}  
\usepackage{caption} 
\usepackage{xcolor}
\usepackage{array} 
\usepackage{booktabs}  
\usepackage{threeparttable}  
\usepackage{multicol}  
\usepackage{multirow}  
\usepackage{natbib}
\usepackage{epsfig}
\usepackage{algorithm}
\usepackage{algorithmic}
\usepackage{amssymb}
\usepackage[utf8]{inputenc}
\usepackage{verbatim}
\usepackage{amsfonts}
\usepackage{amsmath}
\usepackage[switch]{lineno}

\usepackage{multirow}
\usepackage{booktabs}

\usepackage{bm}
\usepackage{subcaption}
\usepackage[utf8]{inputenc}

\usepackage{cleveref}
\crefname{section}{§}{§§}
\Crefname{section}{§}{§§}

\def\ie{\emph{i.e}.~}

\title{Dynamic Hybrid Relation Network for \\
 Cross-Domain Context-Dependent Semantic Parsing}
\author {

        Binyuan Hui,\textsuperscript{\rm 1}
        Ruiying Geng,\textsuperscript{\rm 2}
        Qiyu Ren,\textsuperscript{\rm 3}
        Binhua Li,\textsuperscript{\rm 2}
        Yongbin Li,\textsuperscript{\rm 2\footnote{Corresponding authors: Yongbin Li and Pengfei Zhu}}  \\
        Jian Sun,\textsuperscript{\rm 2}
        Fei Huang,\textsuperscript{\rm 2}
        Luo Si,\textsuperscript{\rm 2}
        Pengfei Zhu,\textsuperscript{\rm 1\footnotemark[1]} 
        Xiaodan Zhu\textsuperscript{\rm 4}
        
}
\affiliations {
    \textsuperscript{\rm 1} College of Intelligence and Computing, Tianjin University, Tianjin, China \\
    \textsuperscript{\rm 2} Alibaba Group, Beijing, China \\
    \textsuperscript{\rm 3} Beijing University of Posts and Telecommunications, Beijing, China \\
    \textsuperscript{\rm 4} Ingenuity Labs Research Institute \& ECE, Queen’s University \\
    \{huibinyuan,zhupengfei\}@tju.edu.cn, {zhu2048}@gmail.com\\ \{ruiying.gry,binhua.lbh,shuide.lyb,jian.sun,f.huang,luo.si\}@alibaba-inc.com \\
     
}
\begin{document}
\maketitle

\begin{abstract}
Semantic parsing has long been a fundamental problem in natural language processing.
Recently, cross-domain context-dependent semantic parsing has become a new focus of research. Central to the problem is the challenge of leveraging contextual information of both natural language utterance and database schemas in the interaction history. 
In this paper, we present a dynamic graph framework that is capable of effectively modelling contextual utterances, tokens, database schemas, and their complicated interaction as the conversation proceeds.
The framework employs a dynamic memory decay mechanism that incorporates inductive bias to integrate enriched contextual relation representation, which is further enhanced with a powerful reranking model.  
At the time of writing, we demonstrate that the proposed framework outperforms all existing models by large margins, achieving new state-of-the-art performance on two large-scale benchmarks, the SParC and CoSQL datasets. Specifically, the model attains a 55.8\% question-match and 30.8\% interaction-match accuracy on SParC, and a 46.8\% question-match and 17.0\% interaction-match accuracy on CoSQL. 
\end{abstract}

\section{Introduction}

Mapping a nature language sentence into a logical form, known as semantic parsing, is a fundamental problem in natural language processing~\cite{DBLP:conf/aaai/ZelleM96,DBLP:conf/uai/ZettlemoyerC05,DBLP:conf/acl/WongM07,DBLP:conf/emnlp/ZettlemoyerC07,Li2014ConstructingAI,Yaghmazadeh2017SQLizerQS,DBLP:conf/acl/IyerKCKZ17}.
Notably, the recent Text-to-SQL tasks have attracted considerable attention, which aims to convert nature language sentences into SQL queries. Large-scale datasets such as SParC \cite{DBLP:conf/acl/YuZYTLLELPCJDPS19} and CoSQL \cite{DBLP:conf/emnlp/YuZELXPLTSLJYSC19} have been made available to train more powerful models. Since relational databases store a great amount of structured data, improving the performance of Text-to-SQL conversion is important for many real-life applications.

\begin{figure}[t]
	\centering
	\includegraphics[width=0.9\columnwidth]{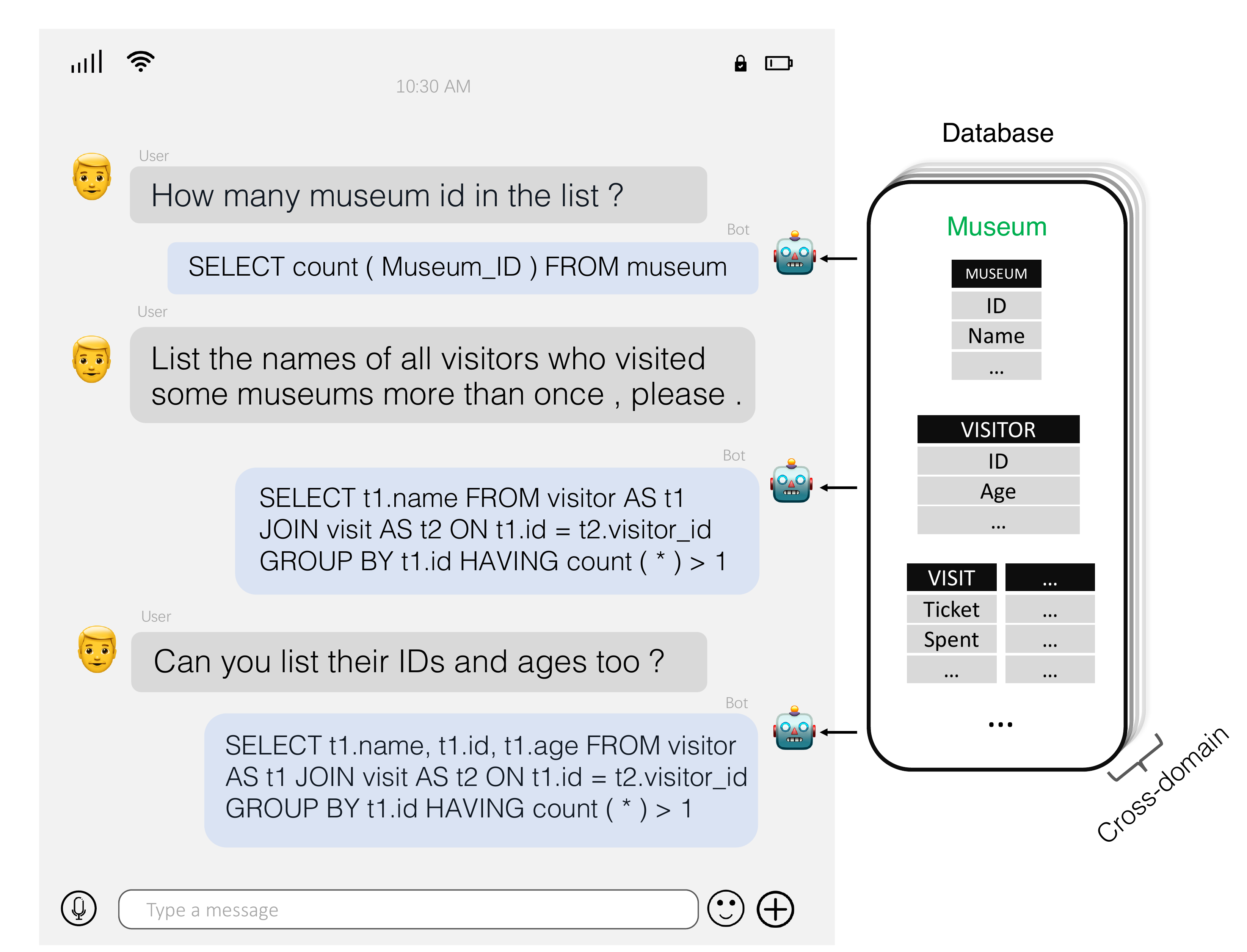}
	\caption{An example of cross-domain context-dependent Text-to-SQL interaction in the CoSQL dataset \cite{DBLP:conf/emnlp/YuZELXPLTSLJYSC19}. }
	\label{Intro}
\end{figure}

Most existing work has focused on precisely converting individual utterances to SQL queries, with a strong assumption on context independence among queries.
However, in real applications as shown in Figure~\ref{Intro}, users interact with cross-domain databases through consecutively communication to exchange information with the databases. 
Unfortunately, the state-of-the-art approaches do not perform well on the newly released, cross-domain context-dependent Text-to-SQL benchmarks, SparC and CoSQL.

Central to the problem of context-dependent semantic parsing is leveraging interactive contextual information of both natural language utterance and database schemas available in the interaction history. Existing work~\cite{DBLP:conf/naacl/SuhrIA18,DBLP:conf/emnlp/ZhangYESXLSXSR19,DBLP:conf/ijcai/LiuCGLZZ20} has mainly focused on integrating context information into the utterance encoding phase. 

In this paper, we present a dynamic graph framework that is capable of effectively modelling contextual utterances, tokens, database schemas, and their complicated interaction when a conversation proceeds.
In our evolving dynamic schema-linking graph network, the representation for nodes, edges, and relation weights adapt in the interactive context. The graph is built at both the utterance and word token level, rendering a flexible framework to model different levels of context in the multi-turn scenario. 
We propose different dynamic memory decay mechanisms to incorporate inductive bias that encourages forgetting the part of dynamic graphs of a longer history, with which dynamic context representation is constructed to leverage both implicit and explicit relations. The framework can effectively leverage features that are powerful in context-independent parsing such as explicit relations \cite{DBLP:conf/acl/WangSLPR20}. We show that such information is also effective for the context-dependent parsing. 

We further design a feature enhanced reranker that integrates external knowledge and task-related representation. The model can identify the correct queries by filtering out those that do not conform to the grammar of SQL and further improve the performance of Text-to-SQL models.

We evaluate our proposed model on two large-scale cross-domain context-dependent benchmarks, SParC \cite{DBLP:conf/acl/YuZYTLLELPCJDPS19} and CoSQL \cite{DBLP:conf/emnlp/YuZELXPLTSLJYSC19}. At the time of writing this paper, the proposed model achieve new state-of-the-art performance on both datasets, substantially outperforming all existing models by large margins. 
Specifically, our model attains a 55.8\% question-match and 30.8\% interaction-match accuracy on SParC, and a 46.8\% question-match and 17.0\% interaction-match accuracy on CoSQL. 
We provide detailed analysis and visualization to further investigate how each component contributes to the entire framework.

\section{Related Work}

\paragraph{Context-independent Semantic Parsing.}
Semantic parsing \cite{DBLP:conf/aaai/ZelleM96,DBLP:conf/uai/ZettlemoyerC05,DBLP:conf/acl/WongM07,DBLP:conf/emnlp/ZettlemoyerC07} maps natural language utterances into logical forms.
Recently, Text-to-SQL is a major focus of semantic parsing in which natural language sentence questioning tables are parsed into SQL queries.
Deep learning has shown to achieve impressive results on context-independent Text-to-SQL datasets such as WikiSQL \cite{DBLP:journals/corr/abs-1709-00103} and Spider \cite{DBLP:conf/emnlp/YuZYYWLMLYRZR18}.
For WikiSQL, \citet{dong-lapata-2018-coarse} propose the Coarse2Fine model which generates meaning sketches abstracted away from  low-level information  such as arguments and variable names and  predicts missing details in order to obtain full meaning representations.
\citet{DBLP:journals/corr/abs-1806-08730} propose MQAN, a model for general question answering that uses a multi-pointer-generator decoder to capitalize on questions as natural language descriptions of tasks.
Furthermore, \citet{Hwang2019ACE} introduce the large pretrained language model and demonstrate the effectiveness of a carefully designed architecture that combines previous approaches.
Compared with Spider, WikiSQL does not involve the complexity of multiple tables.
On Spider, most of work focuses on establishing schema linking, which dynamically obtains the relationships between natural language sentences and database schemas through attention mechanism.
\citet{DBLP:conf/acl/GuoZGXLLZ19} perform schema linking over a question and database schema using customized type vectors for alignment and adopts a grammar-based model \cite{DBLP:conf/acl/YinN17} to synthesize an intermediate representation.
\citet{DBLP:conf/acl/BoginBG19} provide a new perspective on schema linking, which converts the schema to a graph. 
\citet{DBLP:conf/acl/WangSLPR20} propose the RAT-SQL framework, providing a unified way to encode arbitrary relational information among inputs.
Unlike this work, we study context-dependent parsing. 

\begin{figure*}
	\centering
	\includegraphics[width=0.8\linewidth]{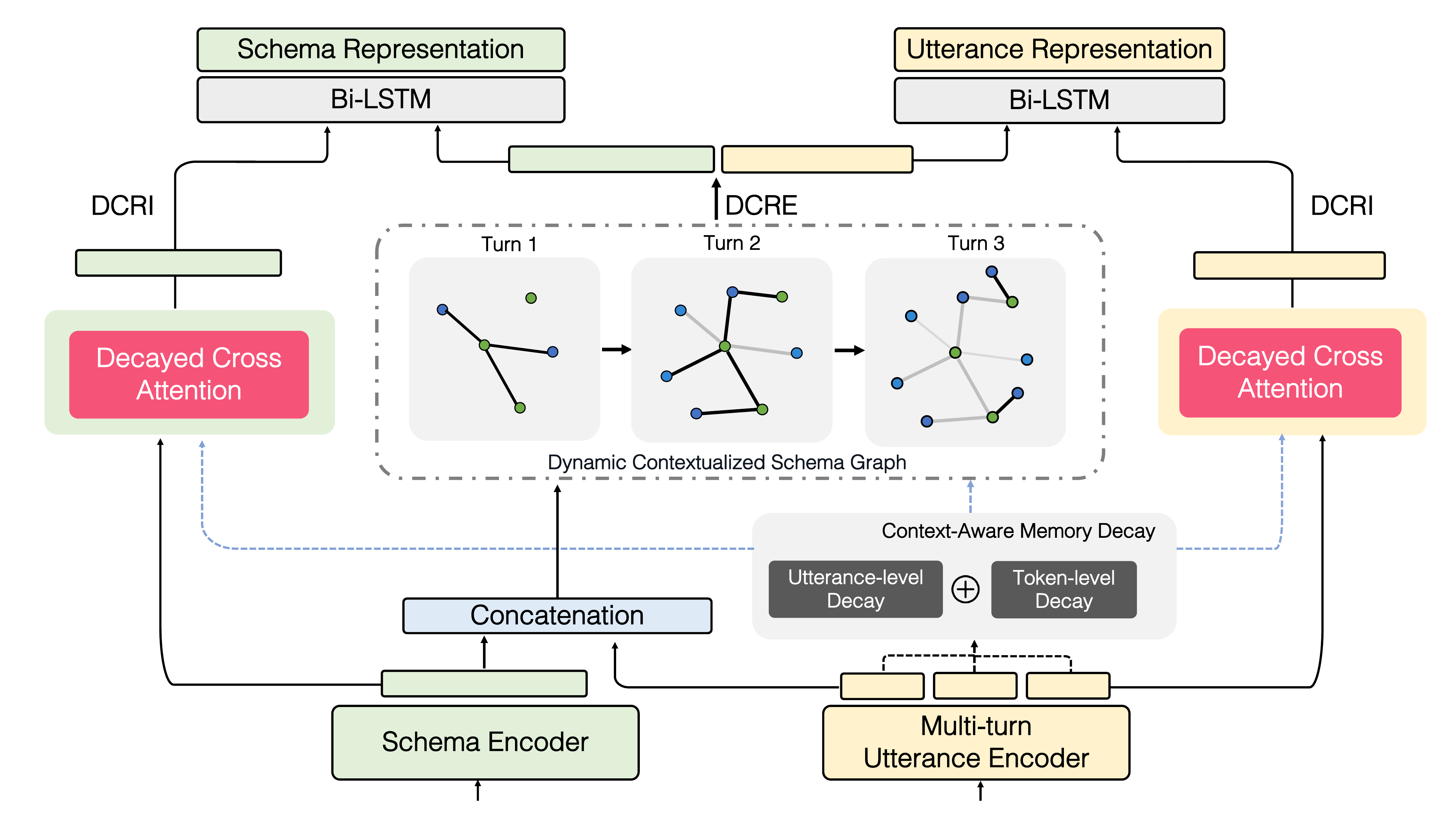}
	\caption{Illustration of the proposed model architecture.}
	\label{arch}
\end{figure*}

\paragraph{Context-dependent Semantic Parsing.}
\citet{DBLP:conf/acl/MillerSBS96} maps utterances to semantic frames, which are then mapped to SQL
queries on the ATIS dataset \cite{DBLP:conf/naacl/HemphillGD90} that has only one database.
Similar to ATIS, SCONE \cite{DBLP:conf/acl/LongPL16,DBLP:conf/acl/GuuPLL17,DBLP:conf/naacl/FriedAK18,DBLP:conf/acl/ArtziS18,DBLP:conf/iclr/HuangCY19} and SequentialQA \cite{DBLP:conf/acl/IyyerYC17} contain no logical form annotations and only denotation \cite{Berant2014SemanticPV} instead.
\citet{DBLP:conf/acl/ZettlemoyerC09} propose a context-independent CCG parser and then applied
it to do context-dependent substitution.
Furthermore, \citet{DBLP:conf/naacl/SuhrIA18} generate ATIS SQL queries from interactions by incorporating history with an interaction-level encoder and copy segments of previously generated queries.
More recently, \citet{DBLP:conf/emnlp/YuZELXPLTSLJYSC19,DBLP:conf/acl/YuZYTLLELPCJDPS19} construct two large-scale cross-domain context-dependent benchmarks for semantic parsing.
Beyond that, \citet{DBLP:conf/emnlp/ZhangYESXLSXSR19} present
an edit-based method that reuses the SQL query generated in the previous time step through editing, which achieves promising results.
\citet{DBLP:conf/ijcai/LiuCGLZZ20} conduct a further exploratory study on semantic parsing in context and perform a fine-grained analysis to explore the sensitivity of input utterance and decoding to context.
As discussed above, we present a dynamic graph framework that is capable of more effectively modelling contextual utterances, tokens, database schemas, and their complicated interaction.

\section{Task Formulation and Notations}
The context-dependent semantic parsing task consists of interactive dialogues in different domains, and its goal is to map nature language utterance in the interaction to the corresponding SQL queries.
Let $\mathcal{I}$ be the set of all interactions, an interaction $I \in \mathcal{I}$ is a series of utterances $\left\langle{x}_{1}, \ldots, {x}_{n}\right\rangle$, and their corresponding SQL queries $\left\langle{y}_{1}, \ldots, {y}_{n}\right\rangle$, where $n$ is the length of the interaction.
In the cross-domain setting, each SQL query is grounded to a multi-table schema and each interaction uses different datasets.
The schema involved in each interaction can be expressed as $S = \left\langle{s}_{1}, \ldots, {s}_{m}\right\rangle$, where $m$ is the number of column headers.
In order to describe the relationships between columns and tables more effectively, for each schema header $s$, the column and table name are formatted as [Table.Column] and [Table.*], respectively.
Given the current utterance $x_i$, the involved schema $S$, and the interaction history of length $i - 1$, formatted as $I[: i-1]=\left\langle\left({x}_{1}, {y}_{1}\right), \ldots,\left({x}_{i-1}, {y}_{i-1}\right)\right\rangle$, the goal is to generate SQL query $y_i$.

\section{Model}
The overall architecture of our proposed model is depicted in Figure~\ref{arch}. In the following sections, we will discuss the components in detail.

\subsection{Encoder}
\label{encoder}
\paragraph{BERT Embedding Input}
The pretrained language models have shown superior performance in many tasks. We utilize BERT \cite{DBLP:conf/naacl/DevlinCLT19} to encode both utterances and schema-related input simultaneously.
Same as in \cite{Hwang2019ACE}, we concatenate all utterances in one conversation and all the schema-related input using \texttt{[SEP]} as the delimiter:
\begin{equation}
[\texttt{CLS}], [x_{1}, \ldots, x_{i}],[\texttt{SEP}], s_{1},[\texttt{SEP}], \ldots, s_{m},[\texttt{SEP}].
\label{input}
\end{equation}
As such, we obtain BERT's utterance and schema representation by feeding the sequence to a pretrained BERT.

\paragraph{Multi-turn Utterance Encoder} To encode the current utterance and effectively integrate information from conversation history, 
at each turn $i$, we employ an utterance-level Bi-LSTM \cite{DBLP:journals/neco/HochreiterS97} to produce embedding from a contextual hidden state:

\begin{equation}
{\mathbf{h}}_{i, k}^{U}=\textbf{Bi-LSTM}^{U}\left(x_{i, k}, {\mathbf{h}}_{i, k-1}^{U}\right) ,
\end{equation}

We use the concatenation of the first and last hidden vector as the utterance encoding.
To take advantage of the utterances in the history, we employ the popular interaction-level encoder \cite{DBLP:conf/naacl/SuhrIA18}. For $i$-th utterance, 
the interaction-level encoder  merges the current utterance embedding $\mathbf{h}_{i}^{U}$ with the preceding interaction-level encoding $\mathbf{h}_{i-1}^{I}$:
\begin{equation}
\mathbf{h}_{i}^{I}=\textbf{LSTM}^{I}\left(\mathbf{h}_{i}^{U}, \mathbf{h}_{i-1}^{I}\right).
\end{equation}
This state is maintained and updated over the entire interaction.
Moreover, we use interaction-level embedding $\mathbf{h}_{i}^{I}$ to further enrich utterance encoding:
\begin{equation}
{\mathbf{h}}_{i, k}^{U}=\textbf{LSTM}^{U}\left([x_{i, k}, \mathbf{h}_{i}^{I}], {\mathbf{h}}_{i, k-1}^{U}\right) .
\end{equation}

\paragraph{Schema Encoder}
To make the model capable of modelling cross-domain information, in addition to utterance encoding, we encode the schema involved in the current interaction.
For each schema input $s$, which consists of table and column names, the schema embedding $\mathbf{h}_{i}^{S}$ are processed by a Bi-LSTM layer:
\begin{equation}
{\mathbf{h}}_{i, k}^{S}=\textbf{Bi-LSTM}^{S}\left(s_{i, k}, {\mathbf{h}}_{i, k-1}^{S}\right).
\end{equation}

\subsection{Dynamic Contextualized Schema Graph}
\label{dynamic graph}

For each conversation that contains multiple turns, we design dynamic contextualized schema graphs, inspired by the recent studies on dynamically evolving structures \cite{pareja2020evolvegcn}. Unlike the original model that was not used for natural language, we construct our dynamic contextualized schema linking graphs, which will change as the conversation proceeds. This graph will be used with dynamic weight decay modules we discuss below to learn enriched contextual representation. 

We would like to jointly learn our representation for utterance $\mathcal{X}$ and schema $\mathcal{S}$ in context $\mathcal{C}$, in particular considering modeling the alignment between them. At the $i$-th turn, given the interaction $X = \{x_{1}^{0 \ldots k_{1}},x_{2}^{0 \ldots k_{2}},\ldots, x_i^{0 \ldots k_{i}}\}$ and the related set of schemas $S = \{{s}_{1}, \ldots, {s}_{m}\}$, we define the \textit{dynamic contextualized schema graph} to be $\mathcal{G}_\mathcal{C}= \langle \mathcal{V}_\mathcal{C}, \mathcal{E}_\mathcal{C} \rangle
$, where $\mathcal{V}_\mathcal{C} = X \cup S$, and $\mathcal{E}_\mathcal{C}$ are schema linking edges among the context words and schema members such as table and column names.
Especially, the relationships can be divided into two categories: 
\begin{itemize}
    \item Internal relations: relations within a database schema, such as a \textit{foreign key}. 
    \item Interactive relations:  relations that align entity references in utterances to the schema columns or tables.
\end{itemize}

\begin{figure}
	\centering
	\includegraphics[width=1\linewidth]{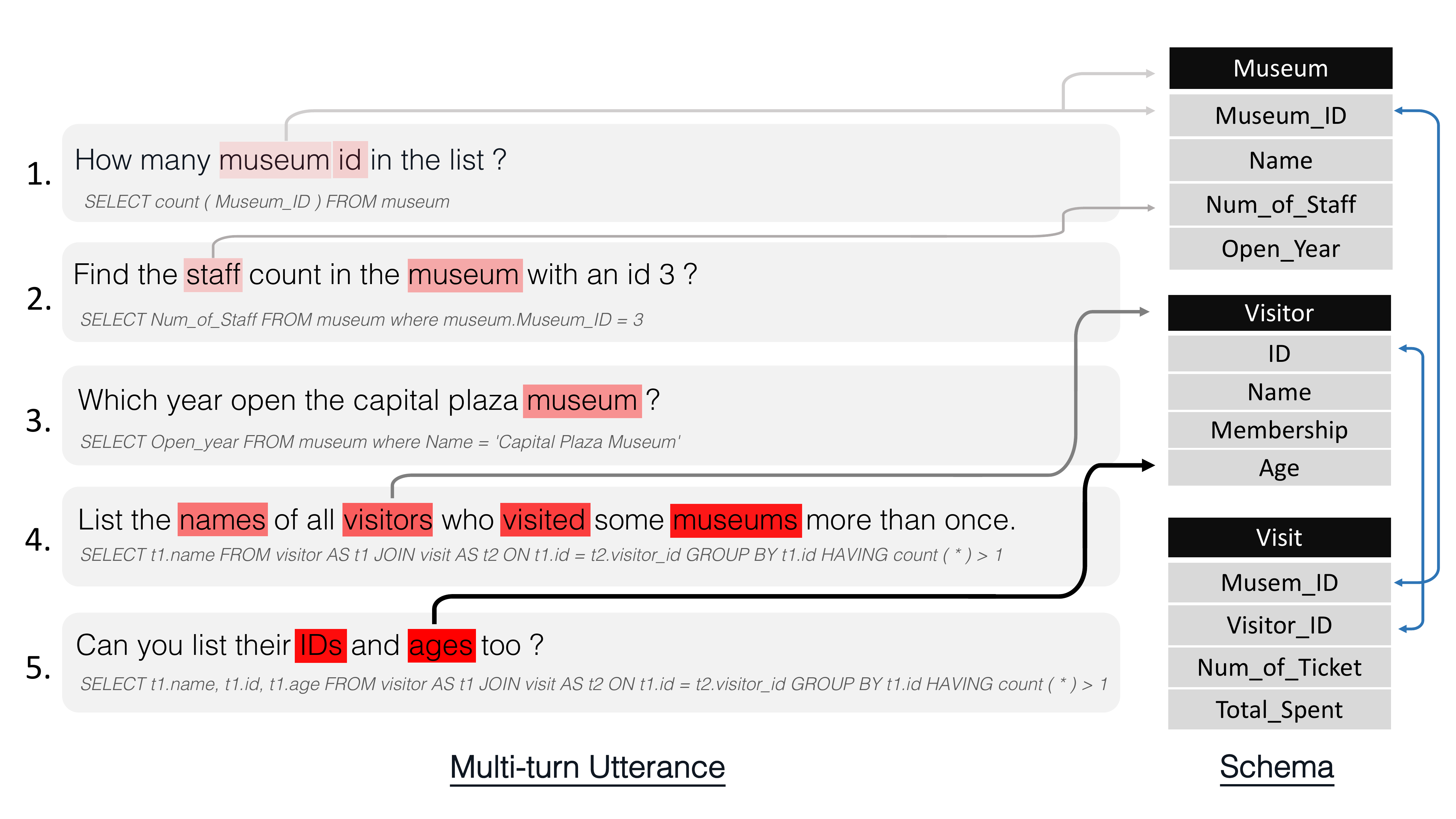}
	\caption{Relations with memory decay. 
	The black line indicates the interactive relation and the blue line indicates the internal relation. The darkness of the color of lines and blocks represents the importance of the links. 
	For simplicity, we did not draw all the linking in the figure.}
	\label{decay}
\end{figure}

\subsection{Context-Aware Memory Decay}
\label{context}
As the conversation proceeds and more queries are asked by users, the contextualized schema graph grows. Note that users' concerns and intention may change frequently. We would like the model to forget unrelated turns that are far away.
In Figure~\ref{decay}, we can see that the first question is about ``Museum", while the second and third are about ``Visit".
We propose to integrate memory decay mechanism to introduce the desired inductive bias. Specifically, we construct both token-level and utterance-level weight decay framework to model the influence of context at different granularities.
For each of the granularities, we provide two approaches: gate-based and schedule-based decay.
\paragraph{Token-level Decay}
We first propose a gate-based decay mechanism to automatically compute the importance of each word token.
The decay weight is computed by:
\begin{equation}
m^T = \mathbf{Sigmoid} (V_{gate} \mathbf{ReLU} (U_{gate} \mathbf{h}^{U}_{i,k})),
\end{equation}
where {$V_{gate}$ and $U_{gate}$} are learned parameters and $\mathbf{h}^U_{i,k}$ are $k$-th token embeddings of the utterance of the $i$-th turn. 
In addition, for schedule-based decay, we investigate three explicit scheduling functions to compute decay weights, borrowed from schedule sampling \cite{bengio2015scheduled}:
\begin{itemize}
    \item Linear decay: $m_t^T = k - c * t$, where $k$ is the base decay weight, $t$ is the position where the decay happens, and $c$ is a constant controlling the slope of decay;
    \item Exponential decay: $m_t^T = k^t$, where $k < 1$;
    \item Inverse sigmoid decay: $m_t^T=  k / (k + \exp^{t / k})$, where $k\geq 1$;
\end{itemize}
We use word tokens' positions in the concatenated sequences (Eq. ~\ref{input}) to compute their distances $t$. 

\paragraph{Utterance-level Decay}
We further propose the utterance-level decay to model the influence of history at the utterance level. Specifically, the gate-decay weight is computed by:
\begin{equation}
m^U = \mathbf{Sigmoid} (V_{gate}^U \mathbf{ReLU} (U_{gate}^U \mathbf{h}^I_i)),
\end{equation}
where $V_{gate}^U$ and $U_{gate}^U$ are learned parameters and $h_i^I$ is the utterance embedding of the $i$-th utterance.  
Similarly, the scheduling function could be used at the utterance-level decay. 
The final memory decay can be empirically selected or by a hyper-parametric weighted combination.

\subsection{Dynamic Context Representation}
\label{representation}
We represent the interaction context based on the dynamic contextualized schema graph $\mathcal{G}_\mathcal{C}$ and memory decay $m$. 

\paragraph{Dynamic Context Representation over Implicit Relations (DCRI)}

We propose a decayed attention mechanism to model implicit relations in the schema graph, which is formulated as $\mathbf{DAttn}(\mathbf{h}^Q, \mathbf{h}^K, m^a)$:
\begin{equation}
\begin{aligned}
&\alpha =\mathbf{softmax}(\mathbf{h}^Q {W}_{\text {att }} \mathbf{h}^K \odot m^a) \\
&\mathbf{DAttn}(\mathbf{h}^Q, \mathbf{h}^K, m^a) =\sum_{l} \alpha_{l} \times \mathbf{h}^K_{l},
\end{aligned}
\end{equation}
where $\mathbf{h}^Q$ and $\mathbf{h}^K$ are query and key embeddings, respectively; ${W}_{\text{att}}$ are learnable parameters and $l$ is the index of key; $m^a$ is the memory decay values used for the current procedure. We split $m$ into $[m^{IU}, m^S]$ to represent the utterance and schema decay values.
First, we utilize the attention mechanism among all headers to explore internal relations in schema, called Schema-Inner Attention module, to update the schema  $\mathbf{h}_{i}^{S}$:
\begin{equation}
\label{inner-att}
\mathbf{h}^{S} = \mathbf{DAttn}(\mathbf{h}^{S}, \mathbf{h}^{S}, m^S).
\end{equation}
It worth noting that the decay value in $m^S$ positions are all set to $1$, since there is no series definition in schema parts.

We then build a decayed attention structure to model the implicit interaction relationships between context utterances and table schemas. We use  $\mathbf{h}^{S}$ to obtain the most relevant columns or tables:

\begin{equation}
\label{co-att}
\begin{aligned}
\mathbf{r}^{IU} &= \mathbf{DAttn}(\mathbf{h}^{U}, \mathbf{h}^{S}, m^{IU}) \\
\mathbf{r}^{IS} &= \mathbf{DAttn}(\mathbf{h}^{S}, \mathbf{h}^{U}, m^{S})
\end{aligned}
\end{equation}
where $\mathbf{r}^{IU}$ and $\mathbf{r}^{IS}$ are implicit exploration representation for utterances and schemas, respectively.

\paragraph{Dynamic Context Representation over Explicit Relations (DCRE)}

Here we further introduce dynamic context representation over the schema graphs with explicit relations, where the relation weights are influenced by the memory decay mechanism discussed above.
First, we concatenate the node embedding of context with that of schema: $\mathbf{h}^{R}= \mathbf{Concat} \left( \mathbf{h}^{U}_{1}; ... ;  \mathbf{h}^{U}_{t};\mathbf{h}^{S}  \right)$ as the input. Then we perform a relation decayed graph transformer to obtain the structured representations:
\begin{equation}
e_{i j}^{(h)}=\frac{\mathbf{h}^{R}_{i} W_{Q}^{(h)}(\mathbf{h}^{R}_{j} {W}_{K}^{(h)} + g_{i j} \odot m_i)^{\top}}{\sqrt{d_{z} / H}}.
\end{equation}
where $W_{Q}^{h}, W_{K}^{h} \in \mathbb{R}^{d_{h}^{R} \times (d_{h}^{R} / H)}$ are learnable parameters, the $H$ is the number of head, $g_{ij}$ is the explicit relationship embedding between the two element $\mathbf{h}^{R}_{i}$ and $\mathbf{h}^{R}_{j}$ from $\mathcal{E}_\mathcal{C}$ in $\mathcal{G}_\mathcal{C}$, and $m_i$ is the memory decay value in the $i$-th position. 

Inspired by \citet{DBLP:conf/acl/WangSLPR20}, we use the following internal relations: (1) \textit{whether columns in the database belong to the same table}; (2) \textit{whether they are foreign keys}.
For the interactive relations, we determine (1) \textit{whether utterance exactly matches the name of column/table}; (2) \textit{whether the n-gram is subsequence of the name of a column/table}. 
We will show that our model can effectively incorporate these features that have been shown to be useful in static graphs in context-independent parsing, and demonstrate they can further improve the performance of our context-independent parsing.

In addition, the attention aggregation operation also needs the decay weights:
\begin{equation}
\begin{aligned}
\alpha_{i j}^{(h)}&=\mathbf{softmax}\left(e_{i j}^{(h)}\right) \\
\mathbf{z}_{i}^{(h)}&=\sum_{j=1}^{n} \alpha_{i j}^{(h)}\left(\mathbf{h}_{j}^{R} W_{V}^{(h)}+g_{i j}  \odot m_i\right). \\
\end{aligned}
\end{equation}
Then we can accumulate the final explicit exploration representation  followed by an $\mathbf{FFN}$ operation \cite{DBLP:conf/nips/VaswaniSPUJGKP17}:

\begin{equation}
\begin{aligned}
\mathbf{z}_{i}&=\mathbf{Concat}\left(\mathbf{z}_{i}^{(1)}, \cdots, \mathbf{z}_{i}^{(H)}\right) \\
{\mathbf{r}}_{i}^{E}&=\mathbf{FFN}(\mathbf{LayerNorm}\left(\boldsymbol{x}_{i}+\boldsymbol{z}_{i}\right)). \\
\end{aligned}
\end{equation}
Finally, we aggregate the embedding from encoder as well as the DCRI and DCRE into new representation for utterance and schema:

\begin{equation}
\begin{aligned}
(\mathbf{r}^{EU}, \mathbf{r}^{ES}) &= \mathbf{Split}(\mathbf{r}^{E}) \\
\mathbf{h}^U &= \textbf{Bi-LSTM}([\mathbf{h}^U, \mathbf{r}^{IU}, \mathbf{r}^{EU}]) \\
\mathbf{h}^S &= \textbf{Bi-LSTM}([\mathbf{h}^S, \mathbf{r}^{IS}, \mathbf{r}^{ES}]) .
\label{hyb}
\end{aligned}
\end{equation}
Here, we consider that DCRE and DCRI establish the schema linking from different perspectives: DCRE pays more attention to the provided prior relationships via exact or n-gram matching. 
Compared to that, DCRI focuses more on semantic relations between utterance and schemas that are not directly captured in surface-form matching.

\subsection{Decoder}
\label{decoder}
We use an LSTM decoder with attention to generate SQL queries at time step $k$:
\begin{equation}
\mathbf{h}_{k}^{D}=\textbf{LSTM}^{D}\left(\left[\mathbf{q}_{k} ; \mathbf{c}_{k}\right], \mathbf{h}_{k-1}^{D}\right),
\end{equation}
where $h^D$ is the hidden state of the decoder and $c_k$ is the context vector with the utterance and schema attention:
\begin{equation}
\begin{aligned}
\mathbf{c}^U = \mathbf{Attn}(\mathbf{h}^{D}, \mathbf{h}^{U}) \\
\mathbf{c}^S = \mathbf{Attn}(\mathbf{h}^{D}, \mathbf{h}^{S}) \\
\mathbf{c} = \mathbf{Concat}(\mathbf{c}^U, \mathbf{c}^S) \\
\end{aligned}
\label{dcontext}
\end{equation}

We apply separate layers to score SQL keywords and column headers and finally use $\mathbf{softmax}$ to generate the output probability distribution:

\begin{equation}
\begin{aligned}
\mathbf{o}_{k} &=\tanh \left(\left[\mathbf{h}_{k}^{D} ; \mathbf{c}_{k}\right] \mathbf{W}_{o}\right) \\
\mathbf{m}^{\mathrm{SQL}} &=\mathbf{o}_{k} \mathbf{W}_{\mathrm{SQL}}+\mathbf{b}_{\mathrm{SQL}} \\
\mathbf{m}^{\text {column }} &=\mathbf{o}_{k} \mathbf{W}_{\mathrm{column}} \mathbf{h}^{S} \\
P\left(y_{k}\right) &=\mathbf{softmax}\left(\left[\mathbf{m}^{\mathrm{SQL}} ; \mathbf{m}^{\mathrm{column}}\right]\right)
\end{aligned}
\end{equation}
In addition, we use query editing mechanism \cite{DBLP:conf/emnlp/ZhangYESXLSXSR19} in the decoder progress to edit the previously generated query while incorporating the context of user utterances and schemas.

\begin{table*}
    \small
	\centering
	\scalebox{0.9}{
	\begin{tabular}{lcccccccc}
	    \toprule
		\textbf{Dataset} & \# sequence & \# user questions & \# databases & \# domain & \# tables & Avg. len & Vocab & Avg. turns  \\
		\midrule
		SParC & 4,298 & 12,726 & 200 & 138 & 1,020 & 8.1 & 3,794 & 3.0 \\
		CoSQL & 3,007 & 15,498 & 200 & 138 & 1,020 & 11.2 & 9,585 & 5.2 \\
		\bottomrule
	\end{tabular}
	}
	\caption{Comparison of the statistics of cross-domain context-dependent Text-to-SQL datasets.}
	\label{statis}
\end{table*}

\begin{table*}[!htbp]
    \small
	\centering
	\scalebox{0.9}{
	\begin{tabular}{ccccccccc}
		\toprule 
		{\multirow{3}*{\textbf{Model}}} & \multicolumn{4}{c}{SParC} & \multicolumn{4}{c}{CoSQL} \\
		\cmidrule{2-9}
		\multirow{2}{*}{} &
		\multicolumn{2}{c}{Question Match.} &
		\multicolumn{2}{c}{Interaction Match.} &
		\multicolumn{2}{c}{Question Match.} &
		\multicolumn{2}{c}{Interaction Match.} \\
		& Dev & Test & Dev & Test & Dev & Test & Dev & Test  \\
		\midrule
		SyntaxSQL-con & 18.5 & 20.2 & 4.3 & 5.2 & 15.1 & 14.1 & 2.7  & 2.2\\
		CD-Seq2Seq & 21.9 & 23.2 & 8.1 & 7.5 & 13.8 & 13.9 & 2.1  & 2.6\\
		EditSQL & 33.0 & - & 16.4 & - & 22.2 & - & 5.8 & -  \\
		RichContext & 41.8 & - & 20.6 & - & 33.5 & - & 9.6 & -\\
		Ours & 42.4 & - & 21.9 & - & 34.5 & - & 11.0 & - \\
		\midrule
		EditSQL + BERT & 47.2 & 47.9 & 29.5 & 25.3 & 39.9 & 40.8 & 12.3  & 13.7\\
		RichContext + BERT & 52.6 & - & 29.9 & - & 41.0 & - & 14.0  & -\\
		Ours + BERT & \textbf{54.1} & \textbf{55.8} ($\uparrow$ 7.9) & \textbf{35.2} & \textbf{30.8} ($\uparrow$ 5.5) & \textbf{45.7} & \textbf{46.8} ($\uparrow$ 6.0) & \textbf{19.5}  & \textbf{17.0} ($\uparrow$ 3.3)\\
		\bottomrule
	\end{tabular}
	}
	\caption{Performance of various methods over questions (question match) and interactions (interaction match) in SParC and CoSQL.}
    \label{result}
\end{table*}

\paragraph{Feature Enhanced Reranker} 
During decoding, we use beam search to generate the N-best list of SQL candidates.
The generated candidate set often contain the correct SQL, but it is not the one with the highest probability.
Specifically we generate an N-best list using vanilla beam search and rerank the generated responses, which has been validated in other semantic parsing tasks \cite{yin-neubig-2019-reranking}.
Furthermore, we design a novel neural reranker module, which integrates the knowledge of the external pretraining models and the task related hidden representation from Eq \ref{hyb}.
Suppose that the current expectation is $x$, the corresponding prediction SQL of the model output is $y'$, and the current task hidden vector is represented as 
\begin{equation}
\mathbf{h}' = \mathbf{MaxPooling}([\mathbf{h}^U, \mathbf{h}^S]). 
\end{equation}
We take the last layer’s hidden state for the first token $[\texttt{CLS}]$ as the deep knowledge $\mathbf{k}$ by feeding the current utterance $x_i$ together with history and prediction $y'$ to the BERT module:
\begin{equation}
[\texttt{CLS}], x_1, [\texttt{SEP}], ..., [\texttt{SEP}], x_i, [\texttt{SEP}], y'.
\end{equation}
Based on that, we combine the task related representation and deep knowledge for joint training:
\begin{equation}
P(B=1 \mid x_{1, ..., i}, y')=\sigma\left(\mathbf{W}_{task}\mathbf{h'}+\mathbf{W}_{knowledge} \mathbf{k}\right)
\end{equation}
where $B$ is the binary class label; $\sigma(\cdot)$ is the sigmoid function; $\mathbf{W}_{task}$ and $\mathbf{W}_{knowledge}$ are learnable weights.
In this way, we can use the reranker module to reorder the generated candidate sets to improve the probability that the prediction is ground truth but not at the top.

\section{Experiment}

\subsection{Setup}
\paragraph{Dataset.} 
We evaluate the performance of the proposed model on two large-scale benchmark datasets, \ie, SParC \cite{DBLP:conf/acl/YuZYTLLELPCJDPS19} and CoSQL \cite{DBLP:conf/emnlp/YuZELXPLTSLJYSC19}.
Table. \ref{statis} summarizes the statistics of SParC and CoSQL. 
Both contain 200 complex databases in 138 different domains.
Compared with SParC, CoSQL has a larger vocabulary and significantly more turns with frequently semantic changes, making it a more challenging dataset.

\paragraph{Evaluation Metrics.}
Following \cite{DBLP:conf/acl/YuZYTLLELPCJDPS19}, we decompose the predicted SQL into clauses such SELECT, WHERE, GROUP BY, and ORDER BY and compute scores for each clause using set matching separately to avoid ordering issues.
On both SParC and CoSQL, we use the two metrics for evaluation:
\textit{Question match} and \textit{Interaction match}.
When all predicted SQL clauses are correct, the exact set matching score is one for question match, and interaction match requires that each predicted SQL in the interaction is correct.

\paragraph{Implementation Details.}
We utilize PyTorch \cite{DBLP:conf/nips/PaszkeGMLBCKLGA19} to implement our proposed model.
For the model without BERT, we initialize word embedding using GloVe \cite{Pennington2014GloveGV} and model parameters from a uniform distribution and set the hidden size as 300 for each $\textbf{LSTM}$ layer.
For the model with BERT, we use Adam \cite{DBLP:journals/corr/KingmaB14} to minimize the token level cross-entropy loss and set the learning rate as 1e-3 on all modules except for the BERT fine-turn stage, for which a learning rate of 1e-5 is used instead.
In particular, we use the pretrained small uncased BERT model with the 768 hidden size.
For the reranker module, we first extract negative samples from the incorrect queries in the training set, and utilize the downsampling strategy to deal with label imbalance to make positive and negative samples balanced.

\begin{figure}
	\centering
	\includegraphics[width=1\linewidth]{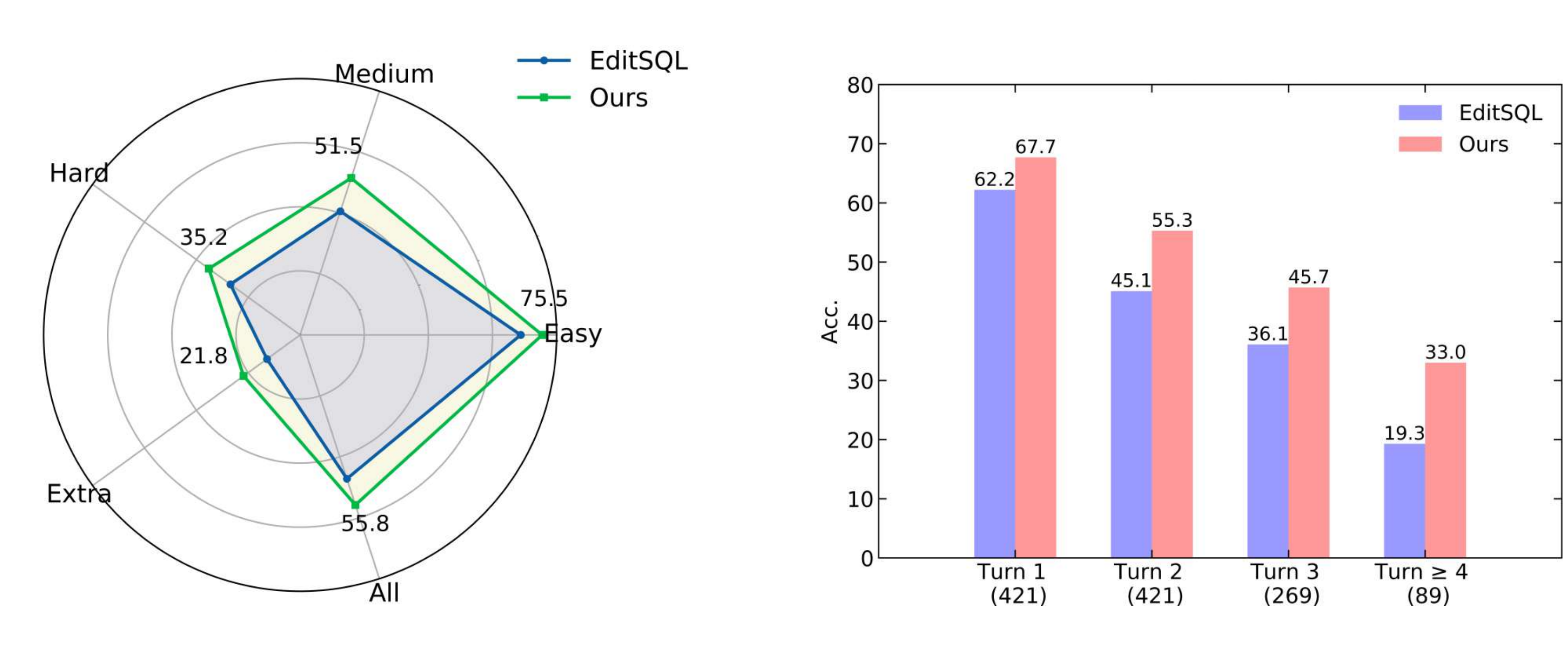}
	\caption{Performance split by different difficulty levels (left) and different turns (right) on SParC.}
	\label{detail}
\end{figure}

\paragraph{Compared Methods.}
We compare the proposed method against the following state-of-the-art models:
\begin{itemize}
    \item \textbf{SyntaxSQL-con} is modified from the original context-independent SyntaxSQLNet \cite{DBLP:conf/emnlp/YuYYZWLR18}, which encodes the utterance and the associated SQL in interaction history. It employs a column attention mechanism to compute representations for the previous question and SQL.
    \item \textbf{CD-Seq2Seq} \cite{DBLP:conf/acl/YuZYTLLELPCJDPS19} is based on sequence-to-sequence modelling extended with the turn-level history encoder proposed in \cite{DBLP:conf/naacl/SuhrIA18}, which modifies the database schema encoder, and takes the bag-of-words representation for column headers as input to perform SQL generation.
    \item \textbf{EditSQL} \cite{DBLP:conf/emnlp/ZhangYESXLSXSR19} is an editing-based encoder-decoder model, which utilizes the interaction history by editing the previous predicted query to improve generation quality. 
    Note that it also uses the implicit relation exploration module based on general attention during the encoding stage.
    \item \textbf{RichContext} \cite{DBLP:conf/ijcai/LiuCGLZZ20} reports state-of-the-art performance on the dev set of SParC and CoSQL and the model is based on rich context modeling methods. We select their best performance to compare with ours. 
\end{itemize}

\subsection{Overall Performance}
As shown in Table \ref{result}, we compare the performance of the proposed model with other state-of-the-art models on the SParC and CoSQL datasets. 
We observe that our method outperforms all existing models on all evaluation metrics. 
Our model achieves the performance of 54.1\%/45.7\% in question match and 35.2\%/19.5\% in interaction match on the dev set, which is a strong model for the context-independent cross-domain Text-to-SQL generation.
We can see that our proposed method outperforms the second published best algorithm (EditSQL) on the SParC / CoSQL test dataset by approximately 7.9\%/5.5\% in question match and 6.0\%/3.3\% in interaction match, presenting new state-of-the-art results on these context-dependent parsing benchmarks. 

In order to distinguish the performance of models of different complexities, following \cite{DBLP:conf/emnlp/YuZYYWLMLYRZR18}, we evaluate the models on the dev dataset with four-levels of difficulty: easy, medium, hard, extra hard. 
Table \ref{detail}(left) shows the comparison between our proposed model and EditSQL, which demonstrates that our method is better than EditSQL.
In particular, in SparC extra hard type, we also achieve a significant improvement. 
Furthermore, to understand how the models perform as the interaction proceeds, Figure \ref{detail}(right) shows the performance split by turns on the SparC dev set.
As the conversation proceeds, it is becomes more difficult to generate SQL using historical information, which leads to a decrease on accuracy.
It shows that although our model is still affected by turns, in SParC, our accuracy rate at the fourth turn is still 33\%, which is close to that of the third turn of the previous method.

\begin{table}[t]
    \small
	\centering
	\scalebox{0.9}{
	\begin{tabular}{lcccccccc}
		\toprule 
		{\multirow{3}*{\textbf{Model}}} & \multicolumn{2}{c}{SParC} & \multicolumn{2}{c}{CoSQL} \\
		\cmidrule{2-5}
		\multirow{2}{*}{} &
		\multicolumn{1}{c}{Ques.} &
		\multicolumn{1}{c}{Int.} &
		\multicolumn{1}{c}{Que.} &
		\multicolumn{1}{c}{Int.} \\
		\midrule
		DCRI & 48.0 & 29.9 & 40.0 & 12.3 \\
		DCRE & 48.3 & 30.2 & 40.6 & 13.0 \\
		DCRI + DCRE & 51.1 & 31.3 & 41.5 & 14.0 \\
		\midrule
		+Gate-based Decay & 51.6 & 31.9 & 42.7 & 15.4 \\
		+Schedule-based Decay & 52.0 & 32.3 & 42.6 & 16.1 \\
		\midrule
		Reranker & 53.7 & 34.6 & 45.0 & 18.6\\
		Feature Enhanced Reranker& \textbf{54.1} & \textbf{35.2} & \textbf{45.7} & \textbf{19.5}\\
		\bottomrule
	\end{tabular}
	}
	\caption{Ablation study of proposed method over Ques. (question match) and Int. (interaction match) in dev set.}
	\label{ablation}
\end{table}

\subsection{Detailed Analysis}
\label{sec:analysis}
\paragraph{Ablation Study.} We investigate the impact of different components on the  performance. 
Specifically, we analyze the following three components:
First, when the relationship is established, two methods are used, Dynamic Context Representation over Implicit Relations (DCRI) and Dynamic Context Representation over Explicit Relations (DCRE). 
Here we perform them without decay to understand the difference.
Second, two methods are used for memory decay: gate-based and schedule-based. 
Third, two types of reranker models, with or without task-related vectors, are further compared and we show that the representation learned from the model is beneficial to the discrimination of the reranker.
As presented in Table \ref{ablation}, we can conclude that each module in our proposed method improve the performance of generation.
It is worth noting that the decay mechanism improves the interaction match of CoSQL more obviously, we attribute this to the more frequent switching of intentions in the CoSQL dataset.

\begin{figure}
	\centering
	\includegraphics[width=1\linewidth]{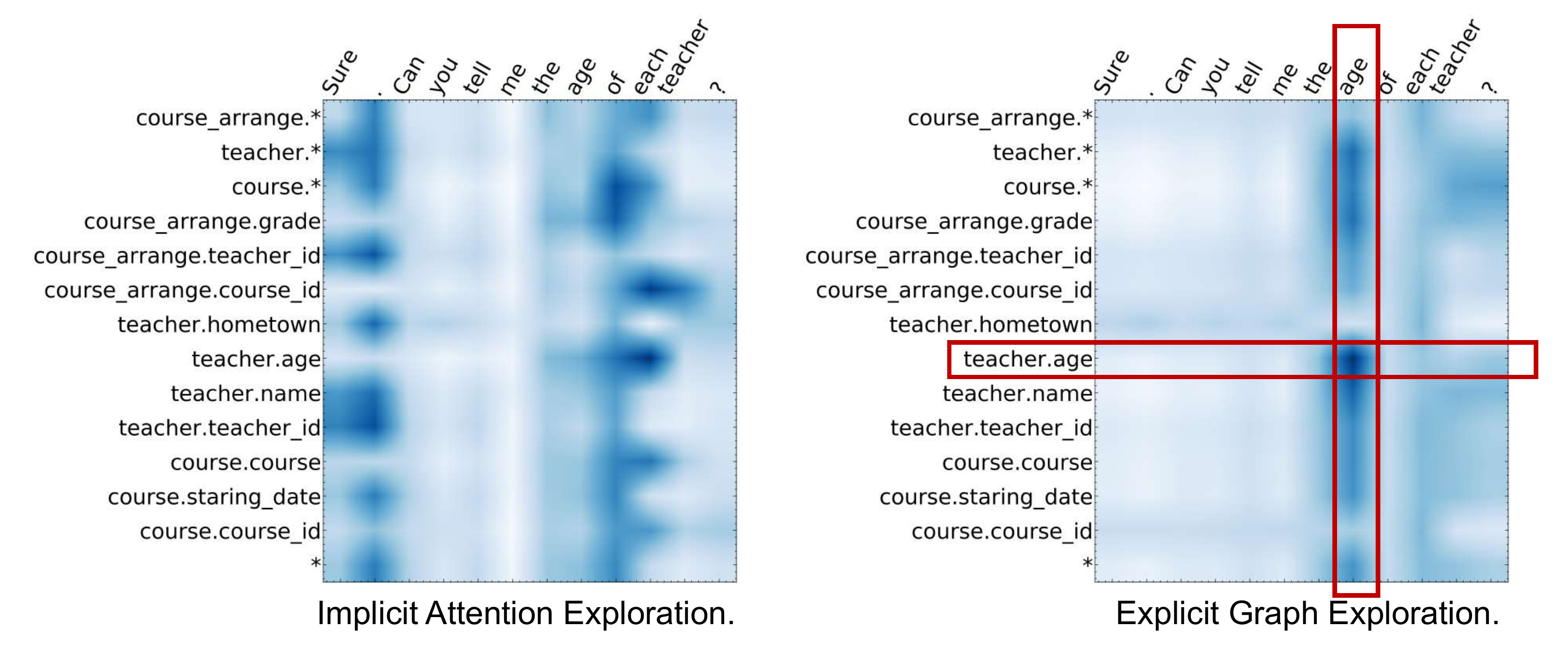}
	\caption{Visualization of DCRI and DCRE in a single turn.}
	\label{diff}
\end{figure}

\begin{figure}
	\centering
	\includegraphics[width=1\linewidth]{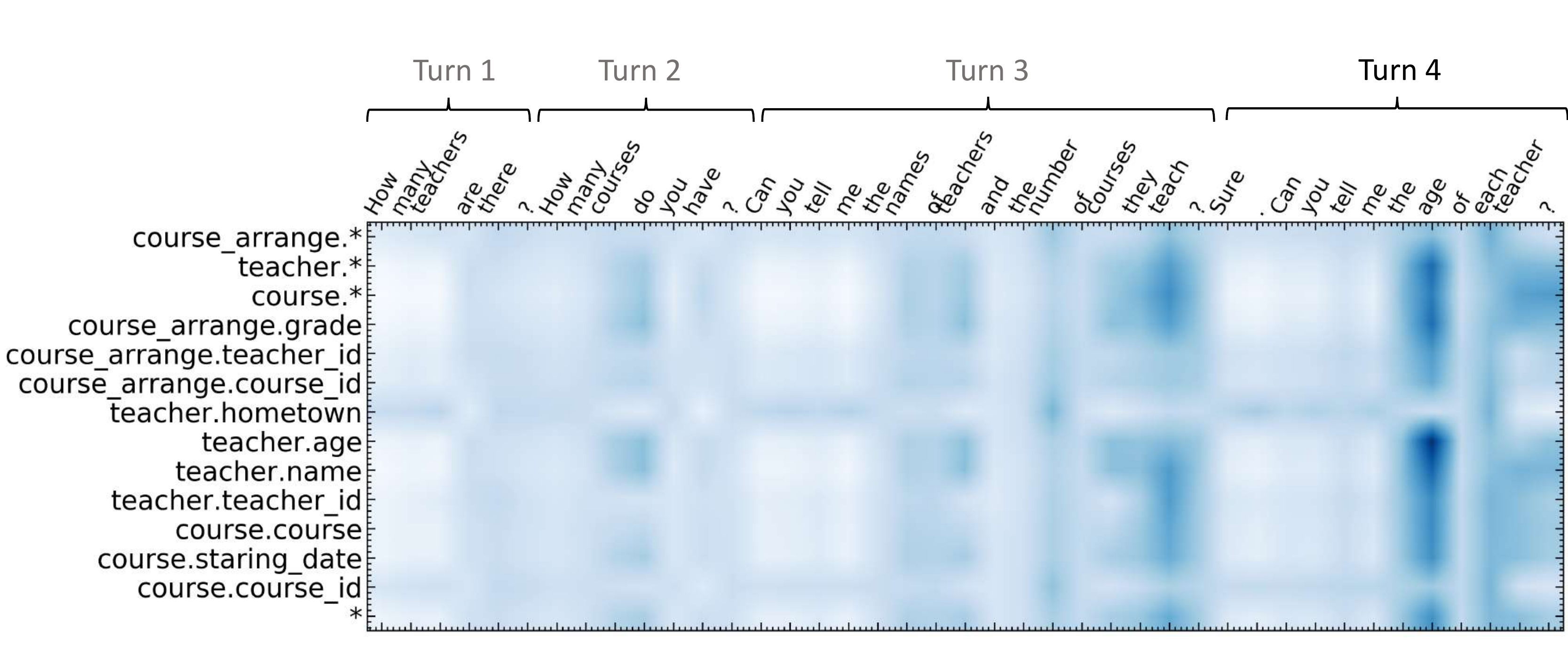}
	\caption{Visualization of DCRE in Context.}
	\label{ex_context}
\end{figure}

\paragraph{Case Study.} In order to further understand the difference between dynamic implicit relation (DCRI) and dynamic explicit relation (DERE), we visualize the relation matrix explored by these two methods. 
As shown in Figure \ref{diff}, the results of implicit exploration that completely relies on attention for learning are not very interpretable, since it provide more high-level semantic representation.
But for explicit graph exploration, we can obtain more interpretable result which benefit from the pre-existing information.
For example, in the red boxes in Figure \ref{diff}, the token \texttt{age} in the utterance has a strong activation with the \texttt{teacher.age} in the schema, which helps establish the link between the utterance and schema.
Furthermore, we visualize the relation of the same case in the context to understand the influence of decay mechanism on the explicit graph.
As shown in Figure \ref{ex_context}, by comparing the activation of different turns, we can observe that with conversation proceeds, the long-term response decreases, and the model pays more attention to the current turn, which is similar to the habit of human memory, which gradually forgets historical information.

\section{Conclusions}
This paper investigates context-dependent semantic parsing. We present a dynamic graph framework that can effectively model contextual utterances, tokens, database schemas, and their complicated relations as the interaction with databases proceeds.
The framework employs dynamic memory decay mechanisms to introduce inductive bias to construct enriched contextual relation representation at both the utterance and token level, rendering a flexible framework to model different levels of context in the dynamic multi-turn scenario. The proposed model achieves new state-of-the-art  performance on two large-scale benchmarks, the SParC and CoSQL datasets. 

\section*{Acknowledgments}
We would like to thank all of the anonymous reviewers for their invaluable suggestions and helpful comments.
We thank Songhe Wang, Tao Yu, Bo Pang for their help with the evaluation on SparC and CoSQL. 
This work was supported by the National Natural Science Foundation of China under Grants 61732011 and 61876127.
The last author's research is supported by NSERC Discovery Grants and Discovery Accelerator Supplements (DAS).

\bibliography{aaai21.bib}

\clearpage
\appendix

\section{Decay Functions}
In Section 4.3, we introduce two decay approaches, including token-level and utterance-level decay.
For each approaches, we also provide two types of functions, \ie, gate-based and schedule based function. For the schedule decay function, we need to set the minimum number $\epsilon$ as $0.8$ to avoid training failure.
We empirically choose the appropriate decay function into model.
In particular, we choose the linear schedule function at the utterance-level on SParC dataset and choose the  inverse sigmoid function at token-level on CoSQL dataset. 
Comparison among different schedule-based functions is shown in the Table \ref{decay_app}.

\begin{table}[!htbp]
	\centering
	\begin{tabular}{lcc}
		\toprule 
		\textbf{Model} & {Ques.} & {Int.} \\
		\midrule
		Linear decay & 42.4 & 15.2  \\
		Exponential decay & 41.9 & 14.3  \\
		Inverse Sigmoid decay & 42.6 & 16.1 \\
		\bottomrule
	\end{tabular}
	\caption{Comparison among different schedule-based functions.}
	\label{decay_app}
\end{table}

\section{Decoder Details}
For the decoder, we exactly follow the previous method of EditSQL’s \cite{DBLP:conf/emnlp/ZhangYESXLSXSR19} decoder component.
The only difference is that we used beam search (the size of beam is $10$) instead of greedy search to generate candidates for reranker.
Here, we description the query editing mechanism of decoder same as EditSQL.
Firstly, the SQL encoder use \textbf{Bi-LSTM} to encode the previous query, and its hidden states are the SQL embeddings.
$\mathbf{c}^{sql}$ is produced by an attention to SQL embeddings.
Then the context vector with the attention to the previous SQL, the Eq. \ref{dcontext} could be rewritten:
\begin{equation}
\begin{aligned}
\mathbf{c} = \mathbf{Concat}(\mathbf{c}^U, \mathbf{c}^S, \mathbf{c}^{sql} ) \\
\end{aligned}
\end{equation}

At each step, the decoder predict a switch $p_{\text {copy}}$ to decide if it need copy from the previous query or insert a new token.
\begin{equation}
\begin{array}{l}
p_{\text {copy }}=\sigma\left(\mathbf{c}_{k} \mathbf{W}_{\text {copy }}+\mathbf{b}_{\text {copy }}\right) \\
p_{\text {insert }}=1-p_{\text {copy }}
\end{array}
\end{equation}
and the output distribution is modified as Eq. \ref{final_d} to take into account the query editing mechanism.

\begin{equation}
\begin{aligned}
& P_{\text {SQL} \cup \text {column }} =\operatorname{softmax}\left(\left[\mathbf{m}^{\text {SQL }} ; \mathbf{~ m}^{\text {column }}\right]\right) \\
& P\left(y_{k}\right)=p_{\text {copy}} \cdot P_{\text {prevSQL }}\left(y_{k} \in \text { prevSQL }\right) \\
& +p_{\text {insert }} \cdot P_{\text {SQL} \cup \text { column }}\left(y_{k} \in \mathrm{SQL} \cup \text { column }\right)
\end{aligned}
\label{final_d}
\end{equation}

\section{Reranker Details}
As the generated SQL queries may not conform to the SQL grammar, the reranker model input is the utterance and generated SQL prediction, the aim is to classificate the current SQL whether matches the current utterance. 
The reranker leverage training data that extract from the intermediate results during downstream model training by BERT model.
More specifically, after training the model for a period of time, when the performance on the entire training set is similar to that on the validation set, we extract negative samples from the incorrect queries in each small batch of beam in the training set, which ensure the distribution of candidates in the training set is similar to that in the validation set. We found this helped us get the best reranking performance. 
For reranker training, we utilize the down sampling strategy to deal with label imbalance by removing some negative examples to make positive and negative samples balanced.
By training the reranker in the training set, the reranker module can decrease the matching scores of inconsistent SQL queries, and increase the consistent ones, improving the final results further in case that the correct query is generated in the beam search.
The reranking module is trained after the main training model finishes, and is used to post-process the output generated in the beam during inference.

\section{Linguistically Oriented Analysis}
In order to discuss the results in a broader sense, we provide some linguistically oriented fine-grained case to show the value that our approach brings.
\citet{DBLP:conf/ijcai/LiuCGLZZ20} performed the linguistically oriented classification for the SParC, and summarized the semantically complete, coreference, ellipsis three types.
Thanks to their classification, we can easily analyze the results.
Table \ref{good} shows some positive results, and we can observe that the proposed model performs well in cases \textit{Demonstrative Pronoun,Possessive Determiner,One Anaphora,Operator}, compared to EditSQL.
Thanks to the Dynamic Contextualized Schema Graph, it is possible to effectively capture contextual information and perform schema linking better, and alleviate problems of coreference and ellipsis.
Unfortunately, the model still performs unsatisfactorily in some cases, such as in Table \ref{bad}, where the \textit{Schema} and \textit{Complex Coreference} problems remain a challenge that can perhaps be corrected in the future by introducing additional knowledge and pre-training the model.

\begin{table*}[!htbp]
    \small
	\centering
	\scalebox{0.75}{
	\begin{tabular}{lll}
		\toprule 
		\textbf{Type} & {Index} & {Case} \\
		\midrule
		\multirow{6}{*}{Demonstrative Pronoun} & $x_1$ & What are all the airlines ? \\
		 & $x_2$ & Of these , which is Jetblue Airways ?  \\
		 & $x_3$ & What is the country corresponding it ?  \\
		 & EditSQL & select Country from airlines where Abbreviation = 1 \\
		 & Ours & select Country from airlines where Airline = 1 \\
		 & GT & select Country from airlines where Airline = 1 \\
		 \hline\hline
		 \multirow{6}{*}{Possessive Determiner} & $x_1$ & What flights land in Aberdeen ? \\
		 & $x_2$ & Also include flights that land in Abilene . \\
		 & $x_3$ & How many are there ? \\
		 & EditSQL & select count ( * ) from airports as T1 join flights as T2 on T1.AirportCode = T2.DestAirport where T1.AirportName = 1 or T1.AirportName = 1 \\
		 & Ours & select count ( * ) from airports as T1 join flights as T2 on T1.AirportCode = T2.DestAirport where T1.City = 1 or T1.City = 1 \\
		 & GT & select count ( * ) from airports as T1 join flights as T2 on T1.AirportCode = T2.DestAirport where T1.City = 1 or T1.City = 1 \\
         \hline\hline
		 \multirow{5}{*}{One Anaphora} & $x_1$ & Order the pets by age	None ? \\
		 & $x_2$ & How much does each one weigh ? \\
		 & EditSQL & select pet\_age , weight from Pets order by pet\_age \\
		 & Ours & select weight from Pets order by pet\_age \\
		 & GT & select weight from Pets order by pet\_age \\
		 \hline\hline
		 \multirow{6}{*}{Operator} & $x_1$ & Find the number of players for each country . \\
		 & $x_2$ & Which country has the least number of players ? Give the country code . \\
		 & $x_3$ & Greatest number ? \\
		 & EditSQL & select country\_code , count ( * ) from players group by country\_code order by count ( * ) desc limit 1 \\
		 & Ours & select country\_code from players group by country\_code order by count ( * ) desc limit 1 \\
		 & GT & select country\_code from players group by country\_code order by count ( * ) desc limit 1 \\
		\bottomrule
	\end{tabular}
	}
	\caption{Positive examples compared to EditSQL.}
	\label{good}
\end{table*}

\begin{table*}[!htbp]
    \small
	\centering
	\scalebox{0.8}{
	\begin{tabular}{lll}
		\toprule 
		\textbf{Type} & {Index} & {Case} \\
		\midrule
		\multirow{4}{*}{Schema} & $x_1$ & How many different losers participated in the Australian Open ? \\
		 & $x_2$ & Winners ?  \\
		 & Ours & select count ( loser\_id ) from matches where tourney\_name = 1 \\
		 & GT & select count(DISTINCT loser\_name) from matches where tourney\_name  =  1 \\
		 \hline\hline
		 \multirow{4}{*}{Complex Coreference} & $x_1$ & What is the maximum population of a country in Asia ? \\
		 & $x_2$ & Which countries in Africa have a population smaller than that ? \\
		 & Ours & select * from country where Population \textless ( select max ( Population ) from country ) \\
		 & GT & select Name from country where Continent  =  "Africa"  and population  \textless  (select max(population) from country where Continent  =  "Asia") \\
		\bottomrule
	\end{tabular}
	}
	\caption{Bad case of the proposed model.}
	\label{bad}
\end{table*}

\end{document}